\documentclass[10pt]{report}

\usepackage{algorithm}
\usepackage{algorithmic}
\usepackage{amsmath, amsthm, amssymb, paralist}
\usepackage{natbib}
\usepackage{graphicx}

\newtheorem{theorem}{Theorem}[section]

\def \x {\mathbf{x}}

\begin{document}

\title{Structure Learning of Probabilistic Graphical Models: A Comprehensive Survey}
\author{Yang Zhou\\Michigan State University}
\date{Nov 2007}
\maketitle

\tableofcontents

\chapter{Graphical Models}
\section{Introduction} 
Probabilistic graphical models combine the graph theory and probability theory to give a multivariate statistical modeling. They provide a unified description of uncertainty using probability and complexity using the graphical model. Especially, graphical models provide the following several useful properties: 
\begin{itemize}
\item Graphical models provide a simple and intuitive interpretation of the structures of probabilistic models. On the other hand, they can be used to design and motivate new models. 

\item Graphical models provide additional insights into the properties of the model, including the conditional independence properties. 

\item Complex computations which are required to perform inference and learning in sophisticated models can be expressed in terms of graphical manipulations, in which the underlying mathematical expressions are carried along implicitly. 
\end{itemize}

The graphical models have been applied to a large number of fields, including bioinformatics, social science, control theory, image processing, marketing analysis, among others. However, structure learning for graphical models remains an open challenge, since one must cope with a combinatorial search over the space of all possible structures. 

In this paper, we present a comprehensive survey of the existing structure learning algorithms. 

\section{Preliminaries} 
We will first define a set of notations which will be used throughout this paper. We represent a graph as $G =\langle V,E
\rangle$ where $V = \{v_i\}$ is the set of nodes in the graph and each node corresponds to a random variable $x_i \in\mathcal{X}$. $E = \{(v_i,v_j): i \neq j\}$ is the set of edges. In a directed graph, if there is an edge $E_{i,j}$ from $v_i$ to $v_j$, then $v_i$ is a parent of node $v_j$ and $v_j$ is a child of node $v_i$. If there is no cycle in a directed graph, we call it a Directed Acyclic Graph (DAG). The number of nodes and number of edges in a graph are denoted by $|V|$ and $|E|$ respectively. $\pi(i)$ is used to represent all the parents of node $v_i$ in a graph. 
$U = \{x_1,\cdots,x_n\}$ denotes the finite set of discrete random variables where each variable $x_i$ may take on values from a finite domain. $Val(x_i)$ denotes the set of values that variable $x_i$ may attain, and $|x_i| = |Val(x_i)|$ denotes the cardinality of this set. In probabilistic graphical network, the Markov blanket $\partial v_i$~\citep{Pearl1988} of a node $v_i$ is defined to be the set of nodes in which each has an edge to $v_i$, i.e., all $v_j$ such that $(v_i,v_j) \in E$. The Markov assumption states that in a probabilistic graphical network, every set of nodes in the network is conditionally independent of $v_i$ when conditioned on its Markov blanket $\partial v_i$. Formally, for distinct nodes $v_i$ and $v_k$, 
\begin{eqnarray*}
P (v_i|\partial v_i \cap v_k)= P (v_i|\partial v_i) 
\end{eqnarray*}
The Markov blanket of a node gives a localized probabilistic interpretation of the node since it identifies all the variables that shield off the node from the rest of the network, which means that the Markov blanket of a node is the only information necessary to predict the behavior of that node. 
A DAG $G$ is an I-Map of a distribution $P$ if all the Markov assumptions implied by $G$ are satisfied by $P$.
\begin{theorem} 
\emph{(Factorization Theorem)}
If $G$ is an I-Map of $P$ , then 
$$
P(x_1,\cdots,x_n)= \prod_i P(x_i|x_{\pi(i)}) 
$$
\end{theorem}

According to this theorem, we can represent $P$ in a compact way when $G$ is sparse such that the number of parameter needed is linear in the number of variables. This theorem is true in the reverse direction. 

The set $X$ is d-separated from set $Y$ given set $Z$ if all paths from a node in $X$ to a node in $Y$ are blocked given $Z$. 

The graphical models can essentially be divided into two groups: directed graphical models and undirected graphical models.

\begin{figure}[!ht]
\centerline{\includegraphics[width=0.25\textwidth]{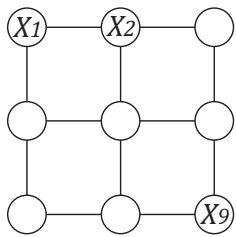}}
\caption{An Ising model with 9 nodes.}\label{fig:ising}
\end{figure}

\section{Undirected Graphical Models} 
\subsection{Markov Random Field}
A Markov Random Field (MRF) is defined as a pair $M =\langle G, \Phi \rangle$. Here $G=\langle V,E \rangle$ represents an undirected graph, where $V = \{V_i\}$ is the set of nodes, each of which corresponds to a random variable in $\mathcal{X}$ ; $E = \{(V_i,V_j): i\neq j\}$ represents the set of undirected edges. The existence of an edge $\{u, v\}$ indicates the dependency of the random variable $u$ and $v$. $\Phi$ is a set of potential functions (also called factors or clique potentials) associated with the maximal cliques in the graph $G$. Each potential function $\phi_c(\cdot)$ has the domain of some clique $c$ in $G$, and is a mapping from possible joint assignments (to the elements of $c$) to non-negative real values. A maximal clique of a graph is a fully connected sub-graph that can not be further extended. We use $C$ to represent the set of maximal cliques in the graph. $\phi_c$ is the potential function for a maximal clique $c \in\mathcal{C}$. The joint probability of a configuration $x$ of the variables $V$ can be calculated as the normalized product of the potential function over all the maximal cliques in $G$:
\begin{eqnarray*}
P(\mathbf{x})=\frac{\prod_{c\in\mathcal{C}}\phi_c(\mathbf{x}_c)}{\sum_{\mathbf{x}'_c}\prod_{c\in\mathcal{C}} \phi_c(\mathbf{x}_c)}
\end{eqnarray*}
where $\mathbf{x}_c$ represents the current configuration of variables in the maximal clique $c$, $\mathbf{x}'_c$ represents any possible configuration of variable in the maximal clique $c$. In practice, a Markov network is often conveniently expressed as a log-linear model, given by 
\begin{eqnarray*}
P(\mathbf{x})=\frac{\exp\left(\sum_{c\in C}w_c\phi_c(\mathbf{x}_c)\right)}{\sum_{\mathbf{x}\in\mathcal{X}}\exp\left(\sum_{c\in\mathcal{C}}w_c\phi_c(\mathbf{x}_c)\right)}
\end{eqnarray*}
In the above equation, $\phi_c$ are feature functions from some subset of $X$ to real values, $w_c$ are weights which are to be determined from training samples. A log-linear model can provide more compact representations for any distributions, especially when the variables have large domains. This representation is also convenient in analysis because its negative log likelihood is convex. However, evaluating the likelihood or gradient of the likelihood of a model requires inference in the model, which is generally computationally intractable due to the difficulty in calculating the partitioning function. 

The Ising model is a special case of Markov Random Field. It comes from statistical physics, where each node represents the spin of a particle. In an Ising model, the graph is a grid, so each edge is a clique. Each node in the Ising model takes binary values $\{0, 1\}$. The parameters are $\theta_i$ representing the external field on particle $i$, and $\theta_{ij}$ representing the attraction between particles 
$i$ and $j$. $\theta_{ij} =0$ if $i$ and $j$ are not adjacent. The probability distribution is: 
\begin{eqnarray*}
p(x|\theta)&=&\exp\left(\sum_{i<j}\theta_{ij}x_i x_j +\sum_i \theta_i x_i =- A(\theta) \right)\\
&=& \frac{1}{Z(\theta)} \exp \left(\sum_{i<j}\theta_{ij}x_i x_j + \sum_i \theta_i x_i \right)
\end{eqnarray*}
where $Z(\theta)$ is the partition function. 

\subsection{Gaussian Graphical Model} 
A \emph{Gaussian Graphical Model} (GGM) models the Gaussian property of multivariate in an undirected graphical topology. Assuming that there are $n$ variables and all variables are normalized so that each of them follows a standard Gaussian distribution. We use $\mathbf{X} =(\mathbf{x}_1,\cdots, \mathbf{x}_n)$ to represent the $n\times1$ column matrix. In a GGM, the variables $\mathbf{X}$ are assumed to follow a multivariate Gaussian distribution with covariance matrix $\Sigma$, 
\begin{eqnarray*}
	P(X)=\frac{1}{(2\pi)^{\frac{n}{2}}|\Sigma|^{\frac{1}{2}}}\exp\left(-\frac{1}{2}\mathbf{X}^{\top}\Sigma^{-1}\mathbf{X} \right)
\end{eqnarray*}

In a Gaussian Graphical Model, the existence of an edge between two nodes indicates that these two nodes are not conditionally independent given other nodes. Matrix $\Omega=\Sigma^{-1}$ is called the \emph{precision matrix}. The non-zeros elements in the precision matrix ½ correspond to the edges in the Gaussian Graphical Model.

\section{Directed Graphical Models} 
The most commonly used directed probabilistic graphical model is Bayesian Network~\citep{Pearl1988}, which is a compact graphical representation of joint distributions. A Bayesian Network exploits the underlying conditional independencies in the domain, and compactly represent a joint distribution over variables by taking advantages of the local conditional independence structures. 
A Bayesian network $\mathcal{B} =\langle G,P \rangle$ is made of two components: a directed acyclic graph (DAG) $G$ whose nodes correspond to the random variables, and a set of \emph{conditional probabilistic distributions} (CPD), $P(x_i|\mathbf{x}_{\pi(i)})$, which describe the statistical relationship between each node $i$ and its parents $\pi(i)$. In a CPD, for any specific configuration of $\mathbf{x}_{\pi(i)}$, the sum over all possible values of $x_i$ is 1, 
\begin{eqnarray*}
\sum_{x_i\in Val(x_i)}P(x_i|\mathbf{x}_{\pi(i)})=1.
\end{eqnarray*}

In the continuous case,
\begin{eqnarray*}
\int_{x_i\in Val(x_i)} P(x_i|\mathbf{x}_{\pi(i)}) \mathrm{d} x_i=1
\end{eqnarray*}

where $P(x_i|\mathbf{x}_{\pi(i)})$ is the conditional density function. 
The conditional independence assumptions together with the CPDs uniquely determine a joint probability distribution via the \emph{chain rule}:
 
\begin{eqnarray*}
P(x_1,\cdots,x_n)=\prod_{i=1}^n P(x_i|\mathbf{x}_{\pi(i)})
\end{eqnarray*}

\begin{figure}[!ht]
\centerline{\includegraphics[width=0.85\textwidth]{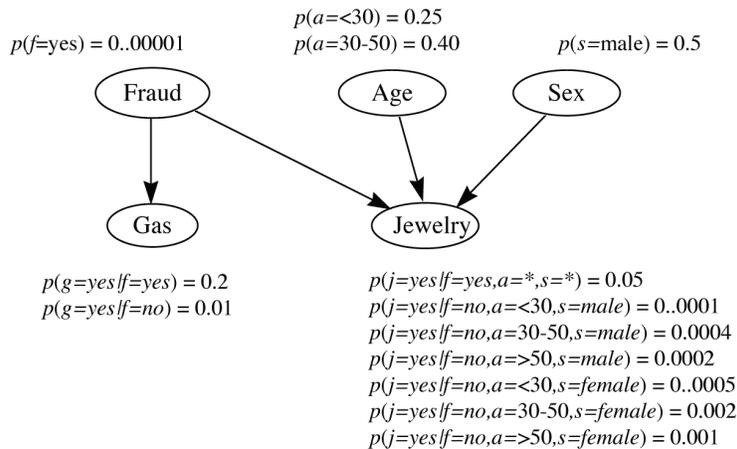}}
\caption{A Bayesian network for detecting credit-card fraud. Arcs indicate the causal relationship. The local conditional probability distributions associated with a node are shown next to the node. The asterisk indicates any value for that variable. Figure excerpted from~\citep{Heckerman1995}.}\label{fig:1_bn}
\end{figure}

\subsection{Conditional Probability Distribution} 
The CPDs may be represented in different ways. The choice of the representation is critical because it specifies the intrinsic nature of the conditional dependencies as well as the number of parameters needed for this representation. Here we describe some different types of CPDs.
 
\subsubsection{Table CPDs} 
In the discrete case, the CPDs can be simply represented as a table in which each row corresponds to a specific configuration of a node and its parents, as well as the corresponding conditional probability of this configuration~\citep{Heckerman1995}. The table CPDs are advantageous in that they are simple and clear, but the size of the table CPDs will grow exponentially with the increase in the number of parents and the number of values that each node can take. 

\subsubsection{Tree CPDs} 
The tree CPDs try to exploit the context specific information (CSI), i.e., the distribution over the values of a node does not depend on the value of some subset of its parents given the value of the other parents~\citep{Boutilier1996}. In a tree CPD, each interior vertex represents the splits on the value of some parent vertices, and each leaf represents a probability conditioned on a specific configuration along the path originated from the root. The tree CPDs usually require a substantially smaller number of parameters than table CPDs when CSI holds in many places of the Bayesian network. 

\subsubsection{Softmax CPDs} 
The softmax CPDs approximates the dependency of a discrete variable $x_i$ on its parents  $x_{\pi(i)}$ by a linear threshold function~\cite{Segal2004}. In this case, the value of each node is determined based on the sum of the contributions of the values of all its parents, i.e., the effect of $\pi(i)$ on node $i$ taking on a value $x_i$ can be summarized via a linear function: 
\begin{eqnarray*}
f_{x_i}(\mathbf{x}_{\pi(i)}=\sum_{j=1}^{|\pi(i)|}w_{x_i,j}x_{\pi(i) (j)}
\end{eqnarray*}

In the above equation, each weight $w_{x_i,j}$ represents the contribution of the jth parent to the value of the target node $i$. Given the contribution function f,a common choice of how the probability of $x_i$ depends on $f_{x_i}(\mathbf{x}_{\pi(i)})$ is the \emph{softmax} distribution, which is the standard extension of the binary logistic conditional distribution to the multi-class case: 
\begin{eqnarray*}
P(x_i|\mathbf{x}_{\pi(i)}) = \frac{\exp(f_{x_i}(\mathbf{x}_{\pi(i)}))}{\sum_{x_i\in Val(x_i)}\exp (f_{x_i} (\mathbf{x}_{\pi(i)}))}
\end{eqnarray*}

\subsubsection{Gaussian CPDs} 
In many cases the random variables are continuous with associated density functions. A common choice of the density function is Gaussian distribution or \emph{Normal distribution} $t\sim N(\mu,\sigma^2)$: 
\begin{eqnarray*}
P(x_i=t) = N(\mu,\sigma^2) = \frac{1}{\sqrt{2\pi }\sigma}\exp\left(\frac{(t-\mu)^2}{2\sigma^2} \right)
\end{eqnarray*}

The Gaussian CPDs are often incorporated in the table or tree CPDs, in which the parameters $mu$ and $\sigma$ of the Gaussian distribution are determined by the configuration of node $i$Õs parents.
 
\subsubsection{Sigmoid CPDs} 
The Sigmoid Belief Networks (SBN)~\citep{Neal1992,Titov2007} has the CPD in the form: 
\begin{eqnarray*}
P(x_i=1|\mathbf{x}_{\pi(i)})=\sigma(\sum_{j\in\pi(i)}J_{ij}x_j)
\end{eqnarray*}
where $\sigma(\cdot)$ denotes the logistic sigmoid function, and $J_{ij}$ is the weight from $j$ to $i$.
 
\subsubsection{Probability Formulas CPDs} 
In a Relational Bayesian Network (RBN)~\citep{Jaeger1997,Jaeger2001}, all variables take binary values. Each root node $i$ has probability $\theta_i\in[0,1]$ to be 1. For each non-root node, the probability to taking value 1 is a combination function of the values of all its parents. A commonly used combination function is the noisy-or function which is defined as $\text{noisy-or}(I)=1 - \pi_{p\in I}(1-p)$ where $I$ is a multiset of probabilities.

\section{Other Graphical Models} 
\begin{itemize}
\item \emph{Dependency Networks}: In~\citep{Heckerman2000}, the authors proposed a probabilistic graphical model named Dependency Networks, which can be considered as combination of Bayesian network and Markov network. The graph of a dependency network, unlike a Bayesian network, can be cyclic. The probability component of a dependency network, like a Bayesian network, is a set of conditional distributions, one for each node given its parents. 

A dependency network is a pair $\langle G,P \rangle$ where $G$ is a cyclic directed graph and $P$ is a set of conditional probability distributions. The parents of nodes $\pi(i)$ of node $i$ correspond to those variables that satisfy 
\begin{eqnarray*}
p(x_i|\mathbf{x}_{\pi(i)}) = p(x_i|\mathbf{x}_{V\setminus i})
\end{eqnarray*}
 
In other words, a dependency network is simply a collection of conditional distributions that are defined and built separately. In a specific context of sparse normal models, these would define a set of separate conditional linear regressions in which $x_i$ is regressed to a small selected subset of other variables, each being determined separately. 

The independencies in a dependency network are the same as those of a Markov network with the same adjacencies. The authors proved that the Gibbs sampler applied to the dependency network will yield a joint distribution for the domain. The applications of dependency network include probabilistic inference, collaborative filtering and the visualization of causal predictive relationships. 

\item \emph{Module Networks}: In~\citep{Segal2003}, the authors proposed a module networks model for gene regulatory network construction. The basic structure is a Bayesian network. Each regulatory module is a set of genes that are regulated in concert by a shared regulation program that governs their behavior. A regulation program specifies the behavior of the genes in the module as a function of the expression level of a small set of regulators. By employing the Bayesian structure learning to the modules instead of genes, this algorithm is able to reduce the computational complexity significantly. 

In~\citep{Toh2002} the authors proposed a model with the similar idea, yet they built a Gaussian Graphical Model instead of Bayesian networks, of module networks. In their study of the yeast (Saccharomyces cerevisiae) genes measured under 79 different conditions, the 2467 genes are first classified into 34 clusters by a hierarchical clustering analysis~\citep{Horimoto2001}. Then the expression levels of the genes in each cluster are averaged for each condition. The averaged expression profile data of 34 clusters were subjected to GGM, and a partial correlation coefficient matrix was obtained as a model of the genetic network. 

\item \emph{Probabilistic Relational Models}: A probabilistic relational model~\citep{Friedman1999a} is a probabilistic description of the relational models, like the models in relational databases. A relational model consists of a set of classes and a set of relations. Each entity type is associated with a set of attributes. Each attribute takes on values in some fixed domain of values. Each relation is typed. The probabilistic relational model describes the relationships between entities and the properties of entities. The model consists of two components: the qualitative dependency structure which is a DAG, and the parameters associated with it. The dependency structure is defined by associating with each attribute and its parents, which is modeled as conditional probabilities. 
\end{itemize}

\section{Network Topology} 
Two classes of network architectures are of special interest to system biology~\citep{Kitano2002}: the \emph{small world networks}~\citep{Watts1998} and \emph{scall-free power law networks}~\citep{Barabasi1999}. Small world networks are characterized by high clustering coefficients and small diameters. The clustering coefficient $C(p)$ is defined as follows. Suppose that a vertex $v$ has $k_v$ neighbors; then at most $k_v(k_v - 1)/2$ edges can exist between them (this occurs when every neighbor of $v$ is connected to every other neighbor of $v$). Let $C_v$ denote the fraction of these allowable edges that actually exist, then the clustering coefficient $C$ is defined as the average of $C_v$ over all $v$. 

These properties reflect the existence of local building blocks together with long-range connectivity. Most nodes in small world networks have approximately the same number of links, and the degree distribution $P (k)$ decays exponentially for large $k$. Compared to small world networks, the scale-free power law networks have smaller clustering coefficients and large diameters. Most nodes in the scale-free networks are connected to a few neighbors, and only a small number of nodes, which is often called ``hubs'', are connected to a large number of nodes. This property is reflected by the power law for the degree distribution $P(k)\sim k^{-v}$. 

Previous studies have found that a number of network structures appear to have structures between the small-world network and the scale-free network. In fact, these networks behave more like \emph{hierarchical scale-free}~\citep{Han2004,Jeong2000,Lukashin2003,Basso2005,Bhan2002,Ravasz2002}. Nodes within the networks are first grouped into modules, whose connectivity is more like the small worlds network. The grouped modules are then connected into a large network, which follows the degree distribution that is similar to that of the scale-free network. 

\section{Structure Learning of Graphical Models} 
There are three major approaches of existing structure learning methods: \emph{constraint-based approaches}, \emph{score-based approaches} and \emph{regression-based} approaches. 

Constraint-based approaches first attempt to identify a set of conditional independence properties, and then attempt to identify the network structure that best satisfies these constraints. The drawback with the constraints based approaches is that it is difficult to reliably identify the conditional independence properties and to optimize the network structure~\citep{Margaritis2003}. Plus, the constraints-based approaches lack an explicit objective function and they do not try to directly find the globally optimal structure. So they do not fit in the probabilistic framework. 

Score-based approaches first define a score function indicating how well the network fits the data, then search through the space of all possible structures to find the one that has the optimal value for the score function. Problem with this approach is that it is intractable to evaluate the score for all structures, so usually heuristics, like greedy search, are used to find the sub-optimal structures. 
Regression-based approaches are gaining popularity in recent years. Algorithms in this category are essentially optimization problems which guarantees global optimum for the objective function, and have better scalability. 

Regression-based approaches are gaining popularity in recent years. Algorithms in this category are essentially optimization problems which guarantees global optimum for the objective function, and have better scalability.

\chapter{Constraint-based Algorithms} 
The constraint-based approaches~\citep{Tsamardinos2006,Juliane2005,Spirtes2000,Wille2004,Margaritis2003,Margaritis1999} employ the conditional independence tests to first identify a set of conditional independence properties, and then attempts to identify the network structure that best satisfies these constraints. The two most popular constraint-based algorithm are the SGS algorithm and PC algorithm~\citep{Tsamardinos2006}, both of which tries to d-separate all the variable pairs with all the possible conditional sets whose sizes are lower than a given threshold. 

One problem with constraint-based approaches is that they are difficult to reliably identify the conditional independence properties and to optimize the network structure~\citep{Margaritis2003}. The constraint-based approaches lack an explicit objective function and they do not try to directly find the global structure with maximum likelihood. So they do not fit in the probabilistic framework. 

\section{The SGS Algorithm}
 
The SGS algorithm (named after Spirtes, Glymour and Scheines) is the most straightforward constraint-based approach for Bayesian network structure learning. It determines the existence of an edge between every two node variables by conducting a number of independence tests between them conditioned on all the possible subsets of other node variables. The pseudo code of the SGS algorithm is listed in Algorithm~\ref{alg:SGS}. After slight modification, SGS algorithm can be used to learn the structure of undirected graphical models (Markov random fields). 
 
\begin{algorithm}[t!] 
\caption{SGS Algorithm} 
\label{alg:SGS}
\begin{algorithmic}[1]
\STATE Build a complete undirected graph $H$ on the vertex set $V$ . 
\STATE 	For each pair of vertices $i$ and $j$, if there exists a subset $S$ of $V\setminus\{i, j\}$ such that $i$ and $j$ are d-separated given $S$, remove the edge between $i$ and $j$ from $G$. 
\STATE	Let $G'$  be the undirected graph resulting from step 2. For each triple of vertices $i$, $j$ and $k$ such that the pair $i$ and $j$ and the pair $j$ and $k$ are each adjacent in $G'$ (written as $i - j - k$) but the pair $i$ and $k$ are not adjacent in $G'$, orient $i - j - k$ as $i \rightarrow j \leftarrow k$ if and only if there is no subset $S$ of $\{j\}\cup V \setminus\{i, j\}$ that d-separate $i$ and $k$. 
\REPEAT 
	\STATE 	If $i \rightarrow j$, $j$ and $k$ are adjacent, $i$ and $k$ are not adjacent, and there is no arrowhead at $j$, then orient $j -k$ as $j \rightarrow k$. 
	\STATE	If there is a directed path from $i$ to $j$, and an edge between $i$ and $j$, then orient $i - j$ as $i \rightarrow j$. 
\UNTIL no more edges can be oriented. 
\end{algorithmic}
\end{algorithm}

The SGS algorithm requires that for each pair of variables adjacent in $G$, all possible subsets of the remaining variables should be conditioned. Thus this algorithm is super-exponential in the graph size (number of vertices) and thus unscalable. The SGS algorithm rapidly becomes infeasible with the increase of the vertices even for sparse graphs. Besides the computational issue, the SGS algorithm has problems of reliability when applied to sample data, because determination of higher order conditional independence relations from sample distribution is generally less reliable than is the determination of lower order independence relations. 

\section{The PC Algorithm} 
The PC algorithm (named after Peter Spirtes and Clark Glymour) is a more efficient constraint-based algorithm. It conducts independence tests between all the variable pairs conditioned on the subsets of other node variables that are sorted by their sizes, from small to large. The subsets whose sizes are larger than a given threshold are not considered. The pseudo-code of the PC algorithm is given in Algorithm~\ref{alg:pc}. We use $\mathcal{N}(i)$ to denote the adjacent vertices to vertex $i$ in a directed acyclic graph $G$.
 
\begin{algorithm}[t!] 
\caption{PC Algorithm}
\label{alg:pc} 
\begin{algorithmic}[1]
\STATE 	Build a complete undirected graph $G$ on the vertex set $V$ . 
\STATE 	n = 0. 
\REPEAT 
	\REPEAT 
		\STATE 	Select an ordered pair of vertices $i$ and $j$ that are adjacent in $G$ such that $\mathcal{N}(i)\setminus\{j\}$ has cardinality greater than or equal to $n$, and a subset $S$ of $\mathcal{N}(i)\setminus\{j\}$ of cardinality $n$, and if $i$ and $j$ are d-separated given $S$ delete edge $i - j$ from $G$ and record $S$ in $Sepset(i, j)$ and $Sepset(j, i)$. 
	\UNTIL all ordered pairs of adjacent variables $i$ and $j$ such that $\mathcal{N}(i)\setminus\{j\}$has cardinality greater than or equal to $n$ and all subsets $S$ of $\mathcal{N}(i)\setminus\{j\}$of cardinality $n$ have been tested for d-separation. 
	\STATE 	n = n + 1. 
\UNTIL for each ordered pair of adjacent vertices $i$ and $j$, $\mathcal{N}(i)\setminus\{j\}$ is of cardinality less than $n$. 
\STATE 	For each triple of vertices $i$, $j$ and $k$ such that the pair $i$, $j$ and the pair $j$, $k$ are each adjacent in $G$ but the pair $i$, $k$ are not adjacent in $G$, orient $i - j - k$ as $i \rightarrow j \leftarrow k$ if and only if $j$ is not in $Sepset(i, k)$. 
\REPEAT 
	\STATE 	If $i \rightarrow j$, $j$ and $k$ are adjacent, $i$ and $k$ are not adjacent, and there is no arrowhead at $j$, then orient $j - k$ as $j \rightarrow k$. 
	\STATE 	If there is a directed path from $i$ to $j$, and an edge between $i$ and $j$, then orient $i - j$ as $i \rightarrow j$. 
\UNTIL no more edges can be oriented. 
\end{algorithmic}
\end{algorithm}

The complexity of the PC algorithm for a graph $G$ is bounded by the largest degree in $G$. Suppose $d$ is the maximal degree of any vertex and $n$ is the number of vertices. In the worst case the number of conditional independence tests required by the PC algorithm is bounded by 
\begin{eqnarray*}
2\binom{n}{2} \sum_{i=1}^d \binom{n-1}{i}
\end{eqnarray*}

The PC algorithm can be applied on graphs with hundreds of nodes. However, it is not scalable if the number of nodes gets even larger. 

\section{The GS Algorithm} 
Both the SGS and PC algorithm start from a complete graph. When the number of nodes in the graph becomes very large, even PC algorithm will be intractable due to the large combinatorial space. 

In~\cite{Margaritis1999}, the authors proposed a Grow-Shrinkage (GS) algorithm to address the large scale network structure learning problem by exploring the sparseness of the graph. The GS algorithm use two phases to estimate a superset of the Markov blanket $\hat{\partial}(j)$ for node $j$ as in Algorithm~\ref{alg:GS1}. In the pseudo code, $i \leftrightarrow_S j$ denotes that node $i$ and $j$ are dependent conditioned on set $S$.
 
\begin{algorithm}[t!] 
\caption{GS: Estimating the Markov Blanket }
\label{alg:GS1} 
\begin{algorithmic}[1] 
\STATE	$S \leftarrow \Phi$. 
\WHILE{$\exists j \in V\setminus\{i\}$ such that $j \leftrightarrow_S i$ }
	\STATE $S \leftarrow S\cup \{j\}$.
\ENDWHILE
\WHILE{$\exists j\in S$ such that $j \nleftrightarrow_{S\setminus\{i\}} i$}
	\STATE $S\leftarrow S\setminus\{j\}$.
\ENDWHILE
\STATE $\hat{\partial}(i)\leftarrow S$
\end{algorithmic}
\end{algorithm}

Algorithm~\ref{alg:GS1} includes two phases to estimate the Markov blanket. In the ``grow'' phase, variables are added to the Markov blanket $\hat{\partial}(j)$ sequentially using a forward feature selection procedure, which often results in a superset of the real Markov blanket. In the ``shrinkage'' phase, variables are deleted from the $\hat{\partial}(j)$ if they are independent from the target variable conditioned on the subset of other variables in $\hat{\partial}(j)$. Given the estimated Markov blanket, the algorithm then tries to identify both the parents and children for each variable as in Algorithm~\ref{alg:GS2}. 
\begin{algorithm}[t!] 
\caption{GS Algorithm}
\label{alg:GS2} 
\begin{algorithmic}[1] 
\STATE 	\emph{Compute Markov Blankets}: for each vertex $i\in V$ compute the Markov blanket $\partial(i)$. 
\STATE 	\emph{Compute Graph Structure}: for all $i\in V$ and $j\in \partial(i)$, determine $j$ to be a direct neighbor of $i$ if $i$ and $j$ are dependent given $S$ for all $S\subseteq T$ where $T$ is the smaller of $\partial(i)\setminus\{j\}$ and $\partial(j)\setminus\{i\}$. 
\STATE	\emph{Orient Edges}: for all $i\in V$ and $j\in \partial(i)$, orient $j\rightarrow i$ if there exists a variable $k\in \partial(i)\setminus\{\partial(j)\cup \{j\}\}$ such that $j$ and $k$ are dependent given  $S\cup\{i\}$ for all $S\subseteq U$ where $U$ is the smaller of $\partial(j)\setminus\{k\}$ and $\partial(k)\setminus\{j\}$. 
\REPEAT 
	\STATE 	Compute the set of edges $C = \{i \rightarrow j \mathrm{such~ that} i \rightarrow j~ \mathrm{is~ part~ of~ a~ cycle}\}$. 
	\STATE	Remove the edge in $C$ that is part of the greatest number of cycles, and put it in $R$. 
\UNTIL there is no cycle exists in the graph. 
\STATE 	\emph{Reverse Edges}: Insert each edge from $R$ in the graph, reversed. 
\STATE	\emph{Propagate Directions}: for all $i\in V$ and $j\in \partial(i)$ such that neither $j\rightarrow i$ nor $i\rightarrow j$, execute the following rule until it no longer applies: if there exists a directed path from $i$ to $j$, orient $i\rightarrow j$. 
\end{algorithmic}
\end{algorithm}

In~\citep{Margaritis1999}, the authors further developed a randomized version of the GS algorithm to handle the situation when 1) the Markov blanket is relatively large, 2) the number of training samples is small compared to the number of variables, or there are noises in the data.

In a sparse network in which the Markov blankets are small, the complexity of GS algorithm is $O(n^2)$ where $n$ is the number of nodes in the graph. Note that GS algorithm can be used to learn undirected graphical structures (Markov Random Fields) after some minor modifications. 

\chapter{Score-based Algorithms} 
Score-based approaches~\citep{Heckerman1995,Friedman1999,Hartemink2001} first posit a criterion by which a given Bayesian network structure can be evaluated on a given dataset, then search through the space of all possible structures and tries to identify the graph with the highest score. Most of the score-based approaches enforce sparsity on the learned structure by penalizing the number of edges in the graph, which leads to a non-convex optimization problem. Score-based approaches are typically based on well established statistical principles such as minimum description length (MDL)~\citep{Lam1994,Friedman1996,Allen2000} or the Bayesian score. The Bayesian scoring approaches was first developed in~\citep{Cooper1992}, and then refined by the BDe score~\citep{Heckerman1995}, which is now one the of best known standards. These scores offer sound and well motivated model selection criteria for Bayesian network structure. The main problem with score based approaches is that their associated optimization problems are intractable. That is, it is NP-hard to compute the optimal Bayesian network structure using Bayesian scores~\citep{Chickering1996a}. Recent researches have shown that for large samples, optimizing Bayesian network structure is NP-hard for all consistent scoring criteria including MDL, BIC and the Bayesian scores~\citep{Chickering2004}. Since the score-based approaches are not scalable for large graphs, they perform searches for the locally optimal solutions in the combinatorial space of structures, and the local optimal solutions they find could be far away from the global optimal solutions, especially in the case when the number of sample configurations is small compared to the number of nodes. 

The space of candidate structures in scoring based approaches is typically restricted to directed models (Bayesian networks) since the computation of typical score metrics involves computing the normalization constant of the graphical model distribution, which is intractable for general undirected models~\citep{Pollard1984}. Estimation of graph structures in undirected models has thus largely been restricted to simple graph classes such as trees~\citep{Chow1968}, poly-trees~\citep{Chow1968} and hypertrees~\citep{Srebro2001}.
 
\section{Score Metrics} 
\subsection{The MDL Score} 
The Minimum Description Length (MDL) principle~\citep{Rissanen1989} aims to minimize the space used to store a model and the data to be encoded in the model. In the case of learning Bayesian network $\mathcal{B}$ which is composed of a graph $G$ and the associated conditional probabilities $P_{\mathcal{B}}$, the MDL criterion requires choosing a network that minimizes the total description length of the network structure and the encoded data, which implies that the learning procedure balances the complexity of the induced network with the degree of accuracy with which the network represents the data. 
 
Since the MDL score of a network is defined as the total description length, it needs to describe the data $U$, the graph structure $G$ and the conditional probability $P$ for a Bayesian network $\mathcal{B} =\langle G,P \rangle$. 
 
To describe $U$, we need to store the number of variables n and the cardinality of each variable $x_i$. We can ignore the description length of $U$ in the total description length since $U$ is the same for all candidate networks.
  
To describe the DAG $G$, we need to store the parents $\pi(i)$ of each variable $x_i$. This description includes the number of parents $|\pi(i)|$ and the index of the set $\pi(i)$ in some enumeration of all  $\binom{n}{|\pi(i)|}$ sets of this cardinality. Since the number of parents $|\pi(i)|$ can be encoded in $\log n$ bits, and the indices of all parents of node $i$ can be encoded in $\log\binom{n}{\pi(i)}$ bits, the description length of the graph structure $G$ is 
\begin{eqnarray}
DL_{graph}(G)=\sum_i \left(\log n + \log \binom{n}{|\pi(i)|} \right)
\label{eq:dl}
\end{eqnarray}
To describe the conditional probability $P$ in the form of CPD, we need to store the parameters in each conditional probability table. The number of parameters used for the table associated with $x_i$ is $|\pi(i)|(|x_i|- 1)$. The description length of these parameters depends on the number of bits used for each numeric parameter. A usual choice is $1/2\log N$~\citep{Friedman1996}. So the description length for $x_i$'s CPD is 
\begin{eqnarray*}
DL_{tab}(x_i)=\frac{1}{2}|\pi(i)|\left(|x_i|-1 \right)\log N
\end{eqnarray*}
To describe the encoding of the training data, we use the probability measure defined by the network $\mathcal{B}$ to construct a Huffman code for the instances in $D$. In this code, the length of each codeword depends on the probability of that instance. According to~\citep{Cover1991}, the optimal encoding length for instance $x_i$ can be approximated as $-\log P_{x_i}$ . So the description length of the data is 
\begin{eqnarray*}
DL_{data}(D|\mathcal{B}) &=& -\sum_{i=1}^N \log P(x_i)\\
	&=& -\sum_i \sum_{x_i, \mathbf{x}_{\pi(i)}}\# (x_i,\mathbf{x}_{\pi(i)})\log P(x_i|\mathbf{x}_{\pi(i)}).
\end{eqnarray*}

In the above equation, $(x_i,\mathbf{x}_{\pi(i)})$ is a local configuration of variable $x_i$ and its parents, $\#(x_i,\mathbf{x}_{\pi(i)})$ is the number of the occurrence of this configuration in the training data. Thus the encoding of the data can be decomposed as the sum of terms that are ``local'' to each CPD, and each term only depends on the counts $\#(x_i,\mathbf{x}_{\pi(i)})$.
 
If $P(x_i|\mathbf{x}_{\pi(i)})$ is represented as a table, then the parameter values that minimize $DL_{data}(D|\mathcal{B})$ are $\theta_{x_i|\mathbf{x}_{\pi(i)}}=\hat{P}(x_i|\mathbf{x}_{\pi(i)})$~\citep{Friedman1998}. If we assign parameters accordingly, then $DL_{data}(D|\mathcal{B})$ can be rewritten in terms of conditional entropy as $N\sum_i H(x_i|\mathbf{x}_{\pi(i)})$, where 
\begin{eqnarray*}
H(X|Y) = -\sum_{x,y}\hat{P}(x,y)\log \hat{P}(x|y)
\end{eqnarray*}
is the conditional entropy of $X$ given $Y$ . The new formula provides an information-theoretic interpretation to the representation of the data: it measures how many bits are necessary to encode the values of $x_i$ once we know $\mathbf{x}_{\pi(i)}$.
 
Finally, by combining the description lengths above, we get the total description length of a Bayesian network as 
\begin{eqnarray}
DL(G,D)=DL_{graph}(G)+\sum_i DL_{tab}(x_i)+N\sum_i H(x_i|\mathbf{x}_{\pi(i)})
\label{eq:3.2}
\end{eqnarray}

\subsection{The BDe Score} 
The Bayesian score for learning Bayesian networks can be derived from methods of Bayesian statistics, one important example of which is BDe score~\citep{Cooper1992,Heckerman1995}. The BDe score is proportional to the posterior probability of the network structure given the data. 
Let $G^h$ denote the hypothesis that the underlying distribution satisfies the independence relations encoded in $G$. Let $\Theta_G$ represent the parameters for the CPDs qualifying $G$. By BayesÕ rule, the posterior probability $P (G^h|D)$ is 
\begin{eqnarray*}
P(G^h|D) = \frac{P(D|G^h)P(G^h)}{P(D)}
\end{eqnarray*}

In the above equation, $1/P(D)$ is the same for all hypothesis, and thus we denote this constant as $\alpha$. The term $P(D|G^h)$ is the probability given the network structure, and $P(G^h)$ is the prior probability of the network structure. They are computed as follows. 

The prior over the network structures is addressed in several literatures. In~\citep{Heckerman1995}, this prior is chosen as $P(G^h)\propto\alpha^{\Delta(G,G')}$, where $\Delta(G,G')$ is the difference in edges between $G$ and a prior network structure $G'$, and $0 <a< 1$ is the penalty for each edge. In~\citep{Friedman1998}, this prior is set as $P(G^h)\propto 2^{-DL_{graph}(G)}$, where $DL_{graph}(G)$ is the description length of the network structure $G$, defined in Equation~\ref{eq:dl}.

The evaluation of $P(D|G^h)$ needs to consider all possible parameter assignments to $G$, namely 
\begin{eqnarray}
P(D|G^h)=\int P(D|\Theta_G,G^h)P(\Theta_G|G^h)\mathrm{d}\Theta_G,
\label{eq:3.3}
\end{eqnarray}
where $P(D|\Theta_G,G^h)$ is the probability of the data given the network structure and parameters. $P(\Theta_G|G^h)$ is the prior probability of the parameters. Under the assumption that each distribution $P(x_i|\mathbf{x}_{\pi(i)})$ can be learned independently of all other distributions~\citep{Heckerman1995}, Equation~\ref{eq:3.3} can be written as 
\begin{eqnarray*}
P(D|G^h)=\prod_i \prod_{\pi(i)}\int \prod_{x_i}\theta_{i,\pi(i)}^{N(x_i,\mathbf{x}_{\pi(i)})}P(\Theta_{i,\pi(i)}|G^h)\mathrm{d}\Theta_{i,\pi(i)}.
\end{eqnarray*}
Note that this decomposition is analogous to the decomposition in Equation~\ref{eq:3.2}. In~\citep{Heckerman1995}, the author suggested that each multinomial distribution $\Theta_{i,\pi(i)}$ takes a \emph{Dirichlet prior}, such that
\begin{eqnarray*}
P(\Theta_{\mathbf{X}})=\beta \prod_x \theta_x^{N'_x}
\end{eqnarray*}
where $N'_x:x\in Val(X)$ is a set of \emph{hyper parameters}, $\beta$ is a normalization constant. Thus, the probability of observing a sequence of values of $X$ with counts $N(x)$ is 
\begin{eqnarray*}
\int \prod_{x}\theta_{x}^{N(x)}P(\Theta_{X}|G^h)\mathrm{d}\Theta_{X}=\frac{\Gamma(\sum_x N'(x))}{\Gamma (\sum_x(N'_x+N(x)))}\prod_x \frac{\Gamma(N'_x+N(x))}{\Gamma(N'_x)}
\end{eqnarray*}
where $\Gamma(x)$ is the \emph{Gamma} function defined as 
\begin{eqnarray*}
\Gamma(x)=\int_0^{\infty}t^{x-t}e^{-t}\mathrm{d}t
\end{eqnarray*}
The Gamma function has the following properties: 
\begin{eqnarray*}
\Gamma(1)&=& 1\\
\Gamma(x+1) &=& x\Gamma(x)
\end{eqnarray*}
If we assign each $\Theta_{i,\pi(i)}$ a Dirichlet prior with hyperparameters $N$ then
\begin{eqnarray*}
P(D|G^h)=\prod_i \prod_{\pi(i)}\frac{\Gamma(\sum_i N'_{i,\pi(i)})}{\Gamma(\sum_i N'_{i,\pi(i)}+N(\pi(i)))}\prod_{x_i}\frac{\Gamma(N'_{i,\pi(i)+N(i,\pi(i))})}{\Gamma(N'_{i,\pi(i)})}
\end{eqnarray*}

\subsection{Bayesian Information Criterion (BIC)} 
A natural criterion that can be used for model selection is the logarithm of the relative posterior probability: 
\begin{eqnarray}
\log P(D,G) = \log P(G) + \log P(D|G)
\label{eq:3.5}
\end{eqnarray}
 
Here the logarithm is used for mathematical convenience. An equivalent criterion that is often used is: 
\begin{eqnarray*}
\log\left(\frac{P(G|D)}{P(G_0|D)} \right)=\log\left(\frac{P(G)}{P(G_0)} \right) + \log\left(\frac{P(D|G)}{P(D|G_0)} \right)
\end{eqnarray*}
The ratio $P(D|G)/P(D|G_0)$ in the above equation is called \emph{BayesÕ factor}~\citep{Kass1995}. Equation~\ref{eq:3.5} consists of two components: the log prior of the structure and the log posterior probability of the structure given the data. In the large-sample approximation we drop the first term. 

Let us examine the second term. It can be expressed by marginalizing all the assignments of the parameters $\Theta$ of the network: 
\begin{eqnarray}
\log P(D|G)=\log \int_{\Theta}P(D|G,\Theta)P(\Theta|G)\mathrm{d}\Theta
\label{eq:3.6}
\end{eqnarray}
In~\citep{Kass1995}, the authors proposed a \emph{Gaussian approximation} for $P(\Theta|D, G) \propto P (D|\Theta,G)P (\Theta|G)$ for large amounts of data. Let 
\begin{eqnarray*}
g(\Theta) \equiv \log(P (D|\Theta,G)P (\Theta|G))
\end{eqnarray*}
 
We assume that $\tilde{\Theta}$ is the \emph{maximum a posteriori} (MAP) configuration of $\Theta$ for $P (\Theta|D, G)$, which also maximizes $g(\Theta)$. Using the second degree Taylor series approximation of $g(\Theta)$ at $\tilde{\Theta}$:
\begin{eqnarray*}
g(\Theta)\approx g(\tilde{\Theta})-\frac{1}{2}(\Theta-\tilde{\Theta})A(\Theta-\tilde{\Theta})^{\top}
\end{eqnarray*}

Where $A$ is the negative Hessian of $g(\Theta)$ at $\tilde{\Theta}$. Thus we get 
\begin{eqnarray}
P(\Theta|D,G) & \propto & P(D|\Theta,G)P(\Theta,G) \nonumber \\
	& \approx & P(D|\tilde{\Theta},S)P(\tilde{\Theta}|S)\exp\left(\frac{1}{2}(\Theta-\tilde{\Theta})A(\Theta-\tilde{\Theta})^{\top} \right)
\label{eq:3.8}
\end{eqnarray}
So we approximate $P (\Theta|D, G)$ as a multivariate Gaussian distribution. Plugging Equation~\ref{eq:3.8} into Equation~\ref{eq:3.6} and we get: 
\begin{eqnarray}
\log P(D|G)\approx \log P(D|\tilde{\Theta},G)+\log P(\tilde{\Theta}|G)+\frac{d}{2}\log(2\pi)-\frac{1}{2}\log |A|
\label{eq:3.9}
\end{eqnarray}
where $d$ is the dimension of $g(\Theta)$. In our case it is the number of free parameters. 

Equation~\ref{eq:3.9} is called a \emph{Laplace approximation}, which is a very accurate approximation with relative error $O(1/N)$ where $N$ is the number of samples in $D$~\citep{Kass1995}. 

However, the computation of $|A|$ is a problem for large-dimension models. We can approximate it using only the diagonal elements of the Hessian $A$, in which case we assume independencies among the parameters. 

In asymptotic analysis, we get a simpler approximation of the Laplace approximation in Equation~\ref{eq:3.9} by retaining only the terms that increase with the number of samples $N$: $\log P(D|\tilde{\Theta},G)$ increases linearly with $N$; $\log |A|$ increases as $d\log N$. And $\tilde{\Theta}$ can be approximated by the maximum likelihood configuration $\tilde{\Theta}$. Thus we get 
\begin{eqnarray}
\log P(D|G)\approx P(D|\tilde{\Theta},S)-\frac{d}{2}\log N
\label{eq:3.10}
\end{eqnarray}

This approximation is called the \emph{Bayesian Information Criterion} (BIC)~\citep{Schwarz1978}. Note that the BIC does not depend on the prior, which means we can use the approximation without assessing a prior. The BIC approximation can be intuitively explained: in Equation~\ref{eq:3.10} , $\log P (D|\tilde{\Theta},G)$ measures how well the parameterized structure predicts the data, and $(d/2 \log N)$ penalizes the complexity of the structure. Compared to the Minimum Description Length score defined in Equation~\ref{eq:3.2}, the BIC score is equivalent to the MDL except term of the description length of the structure.

\section{Search for the Optimal Structure} 
Once the score is defined, the next task is to search in the structure space and find the structure with the highest score. In general, this is an NP-hard problem~\citep{Chickering1996a}. 

Note that one important property of the MDL score or the Bayesian score (when used with a certain class of \emph{factorized} priors such as the BDe priors) is the \emph{decomposability} in presence of complete data, i.e., the scoring functions we discussed earlier can be decomposed in the following way: 
\begin{eqnarray*}
\mathrm{Score}(G:D)=\sum_i \mathrm{Score}(x_i|\mathbf{x}_{\pi(i)}:N_{x_i,\mathbf{x}_{\pi(i)}})
\end{eqnarray*}
where $N_{x_i,\mathbf{x}_{\pi(i)}}$ is the number of occurrences of the configuration $x_i,\mathbf{x}_{\pi(i)}$.
 
The decomposability of the scores is crucial for score-based learning of structures. When searching the possible structures, whenever we make a modification in a local structure, we can readily get the score of the new structure by re-evaluating the score at the modified local structure, while the scores of the rest part of the structure remain unchanged. 

Due to the large space of candidate structures, simple search would inevitably leads to local maxima. To deal with this problem, many algorithms were proposed to constrain the candidate structure space. Here they are listed as follows. 

\subsection{Search over Structure Space}
 
The simplest search algorithm over the structure is the greedy hill-climbing search~\citep{Heckerman1995}. During the hill-climbing search, a series of modifications of the local structures by adding, removing or reversing an edge are made, and the score of the new structure is reevaluated after each modification. The modifications that increase the score in each step is accepted. The pseudo-code of the hill-climbing search for Bayesian network structure learning is listed in Algorithm 5. 

\begin{algorithm}[t!] 
\caption{Hill-climbing search for structure learning } 
\label{alg:5}
\begin{algorithmic}[1]
\STATE Initialize a structure $G'$.
\REPEAT
	\STATE Set $G=G'$.
	\STATE Generate the acyclic graph set $Neighbor(G)$ by adding, removing or reversing an edge in graph $G$.
	\STATE Choose from $Neighbor(G)$ the one with the highest score and assign to $G'$.
\UNTIL{Convergence}.
\end{algorithmic}
\end{algorithm}

Besides the hill-climbing search, many other heuristic searching methods have also been used to learn the structures of Bayesian networks, including the simulated annealing~\citep{Chickering1996}, best-first search \citep{Russel1995} and genetic search \citep{Larranaga1996}.
 
A problem with the generic search procedures is that they do not exploit the knowledge about the expected structure to be learned. As a result, they need to search through a large space of candidate structures. For example, in the hill-climbing structure search in Algorithm~\ref{alg:5}, the size of $Neighbor(G)$ is $O(n^2)$ where $n$ is the number of nodes in the structure. So the algorithm needs to compute the scores of $O(n^2)$ candidate structures in each update (the algorithm also need to check acyclicity of each candidate structure), which renders the algorithm unscalable for large structures.
 
In \citep{Friedman1999}, the authors proposed a Sparse Candidate Hill Climbing (SCHC) algorithm to solve this problem. The SCHC algorithm first estimates the possible candidate parent set for each variable and then use hill-climbing to search in the constrained space. The structure returned by the search can be used in turn to estimate the possible candidate parent set for each variable in the next step. 

The key in SCHC is to estimate the possible parents for each node. Early works \citep{Chow1968,Sahami1996} use \emph{mutual information} to determine if there is an edge between two nodes:
\begin{eqnarray*}
I(X;Y)=\sum_{x,y}\hat{P}(x,y)\log \frac{\hat{P}(x,y)}{\hat{P}(x)\hat{P}(y)}
\end{eqnarray*}

where $\hat{P}(\cdot)$ is the observed frequencies in the dataset. A higher mutual information indicates a stronger dependence between $X$ and $Y$ . Yet this measure is not suitable be used to determine the existence of an edge between two nodes has problems because, for example, it does not consider the information that we already learnt about the structure. Instead, \citet{Friedman1999} proposed two other metrics to evaluate the dependency of two variables. 

\begin{itemize}
\item The first metric is based on an alternative definition of mutual information. The mutual information between $X$ and $Y$ is defined as the distance between the joint distribution of $\hat{P}(X, Y )$ and the distribution $\hat{P}(X)\hat{P}(Y)$, which assumes the independency of the two variables: 
\begin{eqnarray*}
I(X;Y) = D_{KL}\left(\hat{P}(X,Y)||\hat{P}(X)\hat{P}(Y) \right)
\end{eqnarray*}

where $D_{KL}(P||Q)$ is the \emph{Kullback-Leibler divergence} defined as: 
\begin{eqnarray*}
D_{KL}(P(X)||Q(X))=\sum_X P(X)\log \frac{P(X)}{Q(X)}
\end{eqnarray*}
Under this definition, the mutual information measures the error we introduce if we assume the independency of $X$ and $Y$ . During each step of the search process, we already have an estimation of the network B. To utilize this information, similarly, we measure the discrepancy between the estimation $P_{\mathcal{B}}(X, Y )$ and the empirical estimation $\hat{P}(X,Y)$ as: 
\begin{eqnarray*}
D_{KL}(P(X)||Q(X))=\sum_X P(X)\log \frac{P(X)}{Q(X)}
\end{eqnarray*}

One issue with this measure is that it requires to compute $P_{\mathcal{B}}(X_i,Y_i)$ for pairs of variables. When learning networks over large number of variables this can be computationally expensive. However, one can easily approximate these probabilities by using a simple sampling approach. 

\item	The second measure utilizes the Markov property that each node is independent of other nodes given its Markov blanket. First the \emph{conditional mutual information} is defined as: 
\begin{eqnarray*}
I(X;Y|Z)=\sum_Z\hat{P}(Z)D_{KL}(\hat{P}(X,Y|Z)||\hat{P}(X|Z)\hat{P}(Y|Z)).
\end{eqnarray*}
This metric measures the error that is introduced when assuming the conditional independence of $X$ and $Y$ given $Z$. Based upon this, another metric is defined as: 
\begin{eqnarray*}
M_{shield}(X_i,X_j|\mathcal{B})=I(X_i;X_j|X_{\pi(i)})
\end{eqnarray*}

\end{itemize}
 
Note that using either of these two metrics for searching, at the beginning of the search, i.e., $\mathcal{B}_0$ is an empty network, the measure is equivalent to $I(X;Y)$. Later iterations will incorporate the already estimated network structure in choosing the candidate parents.
 
Another problem with the hill-climbing algorithm is the stopping criteria for the search. There are usually two types of stopping criteria: 
\begin{itemize}
\item \emph{Score-based criterion}: the search process terminates when $Score(\mathcal{B}_t)= Score(\mathcal{B}_{t-1})$. In other words, the score of the network can no longer be increased by updating the network from candidate network space. 
	
\item \emph{Candidate-based criterion}: the search process terminates when $C_i^t=C_i^{t-1}$ for all $i$, that is, the candidate space of the network remains unchanged. 
\end{itemize}

Since the score is a monotonically increasing bounded function, the score-based criterion is guaranteed to stop. The candidate-based criterion might enter a loop with no ending, in which case certain heuristics are needed to stop the search. 

There are four problems with the SCHC algorithm. First, the estimation of the candidate sets is not sound (i.e., may not identify the true set of parents), and it may take a number of iterations to converge to an acceptable approximation of the true set of parents. Second, the algorithm needs a pre-defined parameter $k$, the maximum number of parents allowed for any node in the network. If $k$ is underestimated, there is a risk of discovering a suboptimal network. On the other hand, if $k$ is overestimated, the algorithm will include unnecessary parents in the search space, thus jeopardizing the efficiency of the algorithm. Third, as already implied above, the parameter $k$ imposes a uniform sparseness constraint on the network, thus may sacrifice either efficiency or quality of the algorithm. A more efficient way to constrain the search space is the Max-Min Hill-Climbing (MMHC) algorithm  \citep{Tsamardinos2006}, a hybrid algorithm which will be explained in Section~\ref{sec:5.1}. The last problem is that the constraint of the maximum number of parents $k$ will conflict with the scale-free networks due to the existence of hubs (this problem exists for any algorithm that imposes this constraint). 

Using the SCHC search, the number of candidate structures in each update is reduced from $O(n^2)$ to $O(n)$ where $n$ is the number of nodes in the structure. Thus, the algorithm is capable to learn large-scale structures with hundreds of nodes. 

The hill-climbing search is usually applied with multiple restarts and tabu list~\citep{Cvijovicacute1995}. Multiple restarts are used to avoid local optima, and the tabu list is used to record the path of the search so as to avoid loops and local minima.
 
To solve the problem of large candidate structure space and local optima, some other algorithms are proposed as listed in the following.
\begin{itemize} 
\item In~\citep{Moore2003}, the authors proposed a search strategy based on a more complex search operator called optimal reinsertion. In each optimal reinsertion, a target node in the graph is picked and all arcs entering or exiting the target are deleted. Then a globally optimal combination of in-arcs and out-arcs are found and reinserted into the graph subject to some constraints. With the optimal reinsertion operation defined, the search algorithm generates a random ordering of the nodes and applies the operation to each node in the ordering in turn. This procedure is iterated, each with a newly randomized ordering, until no change is made in a full pass. Finally, a conventional hill-climbing is performed to relax the constraint of max number of parents in the optimal reinsertion operator. 
 	
\item In~\citep{Xiang1997}, the authors state that with a class of domain models of probabilistic dependency network, the optimal structure can not be learned through the search procedures that modify a network structure one link at a time. For example, given the $XOR$ nodes there is no benefit in adding any \emph{one} parent individually without the others and so single-link hill-climbing can make no meaningful progress. They propose a multi-link lookahead search for finding decomposable Markov Networks (DMN). This algorithm iterates over a number of levels where at level $i$, the current network is continually modified by the best set of $i$ links until the entropy decrement fails to be significant. 

\item Some algorithms identify the Markov blanket or parent sets by either using conditional independency test, mutual information or regression, then use hill-climbing search over this constrained candidate structure space~\citep{Tsamardinos2006,Schmidt2007}. These algorithms belong to the hybrid methods. Some of them are listed in Section~\ref{sec:5.1}. 
\end{itemize}

\subsection{Search over Ordering Space} 
The acyclicity of the Bayesian network implies an \emph{ordering} property of the structure such that if we order the variables as $\langle x_1,\cdots,x_n \rangle$, each node $x_i$ would have parents only from the set  $\{x_1,\cdots,x_{i-1}\}$. Fundamental observations~\citep{Buntine1991,Cooper1992} have shown that given an ordering on the variables in the network, finding the highest-scoring network consistent with the ordering is not NP-hard. Indeed, if the in-degree of each node is bounded to $k$ and all structures are assumed to have equal probability, then this task can be accomplished in time $O(n^k)$ where $n$ is the number of nodes in the structure. 

Search over the ordering space has some useful properties. First, the ordering space is significantly smaller than the structure space: $2^{O(n \log n)}$ orderings versus $2^{\Omega(n^2)}$ structures where $n$ is the number of nodes in the structure~\citep{Robinson1973}. Second, each update in the ordering search makes a more global modification to the current hypothesis and thus has more chance to avoid local minima. Third, since the acyclicity is already implied in the ordering, there is no need to perform acyclicity checks, which is potentially a costly operation for large networks. 

The main disadvantage of ordering-based search is the need to compute a large set of sufficient statistics ahead of time for each variable and each possible parent set. In the discrete case, these statistics are simply the frequency counts of instantiations: $\#(x_i, \mathbf{x}_{\pi(i)})$ for each $x_i\in Val(x_i)$ and $\mathbf{x}_{\pi(i)} \in Val(\mathbf{x}_{\pi(i)})$. This cost would be very high if the number of samples in the dataset is large. However, the cost can be reduced by using AD-tree data structure~\citep{Moore1998}, or by pruning out possible parents for each node using SCHC~\citep{Friedman1999}, or by sampling a subset of the dataset randomly. 

Here some algorithms that search through the ordering space are listed: 

\begin{itemize}
\item The ordering-based search was first proposed in~\citep{Larranaga1996} which uses a genetic algorithm search over the structures, and thus is very complex and not applicable in practice. 

\item In~\citep{Friedman2003}, the authors proposed to estimate the probability of a structural feature (i.e., an edge) over the set of all orderings by using a Markov Chain Monte Carlo (MCMC) algorithm to sample over the possible orderings. The authors asserts that in the empirical study, different runs of MCMC over network structure typically lead to very different estimates in the posterior probabilities over network structure features, illustrating poor convergence to the stationary distribution. By contrast, different runs of MCMC over orderings converge reliably to the same estimates. 

\item In~\citep{Teyssier2005}, the authors proposed a simple greedy local hill-climbing with random restarts and a tabu list. First the score of an ordering is defined as the score of the best network consistent with it. The algorithm starts with a random ordering of the variables. In each iteration, a swap operation is performed on any two adjacent variables in the ordering. Thus the branching factor for this swap operation is $O(n)$. The search stops at a local maximum when the ordering with the highest score is found. The tabu list is used to prevent the algorithm from reversing a swap that was executed recently in the search. Given an ordering, the algorithm then tries to find the best set of parents for each node using the Sparse Candidate algorithm followed by exhaustive search. 

\item In~\citep{Koivisto2004,Singh2005}, the authors proposed to use Dynamic Programming to search for the optimal structure. The key in the dynamic programming approach is the ordering $\prec$, and the marginal posterior probability of the feature $f$ 
\begin{eqnarray*}
p(f|\prec)=\sum_{\prec}p(\prec|x)p(f|x,\prec)
\end{eqnarray*}

Unlike~\citep{Friedman2003} which uses MCMC to approximate the above value, the dynamic programming approach does exact summation using the permutation tree. Although this approach may find the exactly best structure, the complexity is $O(n2^n + n^{k+1}C(m))$ where $n$ is the number of variables, $k$ is a constant in-degree, and $C(m)$ is the cost of computing a single local marginal conditional likelihood for m data instances. The authors acknowledge that the algorithm is feasible only for $n \leq 26$. 
\end{itemize}\chapter{Regression-based Algorithms} 
\section{Regression Model} 
Given $N$ sample data points as $(x_i,y_i)$ and pre-defined basis functions $\phi(\cdot)$, the task of regression is to find a set of weights $\mathbf{w}$ such that the basis functions give the best prediction of the label $y_i$ from the input $x_i$. The performance of the prediction is given by an loss function $E_D(\mathbf{w})$. For example, in a linear regression, 
\begin{eqnarray}
E_D({\mathbf{w}})=\frac{1}{2}\sum_{i=1}^N \left(y_i-\mathbf{w}^{\top}\phi(\mathbf{x}_i)\right)^2
\label{eq:4.1}
\end{eqnarray}
To avoid over-fitting, a regularizer is usually added to penalize the weights $\mathbf{w}$. So the regularized loss function is: 
\begin{eqnarray}
E(\mathbf{w})=E_D(\mathbf{w})+\lambda E_W(\mathbf{w})
\label{eq:4.2}
\end{eqnarray}
 
The regularizer penalizes each element of $\mathbf{w}$: 
\begin{eqnarray*}
E_W(\mathbf{w}) = \sum_{j=1}^M \alpha_i \|w_j\|_q
\end{eqnarray*}
When all $\alpha_i$'s are the same, then 
\begin{eqnarray*}
E_W(\mathbf{w}) = \|\mathbf{w}_j\|_q
\end{eqnarray*}
 
where $\|\cdot\|_q$ is the $L_q$ norm, $\lambda$ is the regularization coefficient that controls the relative importance of the data-dependent error and the regularization term. With different values of $q$, the regularization term may give different results: 

\begin{enumerate}
\item When $q = 2$, the regularizer is in the form of sum-of-squares 
\begin{eqnarray*}
E_W(\mathbf{w}) = \frac{1}{2}\mathbf{w}^{\top}\mathbf{w}
\end{eqnarray*}
This particular choice of regularizer is known in machine learning literature as \emph{weight decay} \citep{Bishop2006} because in sequential learning algorithm, it encourages weight values to decay towards zeros, unless supported by the data. In statistics, it provides an example of a parameter shrinkage method because it shrinks parameter values towards zero. 

One advantage of the $L_2$ regularizer is that it is rotationally invariant in the feature space. To be specific, given a deterministic learning algorithm $L$, it is rotationally invariant if, for any training set $S$, rotational matrix $M$ and test example $x$, there is $L[S](x)= L[MS](Mx)$. More generally, if $L$ is a stochastic learning algorithm so that its predictions are random, it is rotationally invariant if, for any $S$, $M$ and $x$, the prediction $L[S](x)$ and $L[MS](Mx)$ have the same distribution. A complete proof in the case of logistic regression is given in~\citep{Ng2004}.
 
This quadratic ($L_2$) regularizer is convex, so if the loss function being optimized is also a convex function of the weights, then the regularized loss has a single global optimum. Moreover, if the loss function $E_D(\mathbf{w})$ is in quadratic form, then the minimizer of the total error function has a closed form solution. Specifically, if the data-dependent error $E_D(\mathbf{w})$ is the sum-of-squares error as in Equation~\ref{eq:4.2}, then setting the gradient with respect to $w$ to zero, then the solution is 
\begin{eqnarray*}
\mathbf{w}=(\lambda \mathbf{I}+\Phi^{\top}\Phi)^{-1}\Phi^{\top}\mathbf{t}
\end{eqnarray*}
 
This regularizer is seen in \emph{ridge regression}~\citep{Hoerl2000}, the support vector machine~\citep{Hoerl2000,Schoelkopf2002} and regularization networks~\citep{Girosi1995}.
  
\item 	$q = 1$ is called \emph{lasso} regression in statistics ~\citep{Tibshirani1996}. It has the property that if $\lambda$ is sufficiently large, then some of the coefficients $w_i$ are driven to zero, which leads to a sparse model in which the corresponding basis functions play no role. To see this, note that the minimization of Equation~\ref{eq:4.2} is equivalent to minimizing the unregularized sum-of-squares error subject to the constraint over the parameters: 

\begin{eqnarray}
\label{eq:4.3}
& \mathop{\arg\min} \limits_{\mathbf{w}} & \frac{1}{2}\sum_{i=1}^N \left( y_i - \mathbf{w}^{\top}\phi(\mathbf{x}_i) \right)^2\\
& \mbox{s. t.} & \sum_{j=1}^M \|w_j\|_q\leq \eta
\end{eqnarray}

\begin{figure}[!ht]
\centerline{\includegraphics[width=0.75\textwidth]{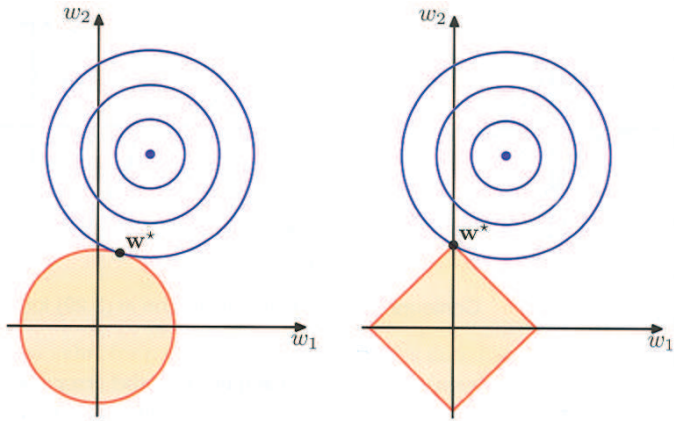}}
\caption{Contours of the unregularized objective function (blue) along with the constraint region (yellow) with $L_2$-regularizer (left) and $L_1$-regularizer. The lasso regression gives a sparse solution. Figure excerpted from~\citep{Bishop2006}.}\label{fig:4.2}
\end{figure} 
 
The Lagrangian of Equation~\ref{eq:4.3} gives Equation \ref{eq:4.2}. The sparsity of the solution can be seen from Figure~\ref{fig:4.2}. Theoretical study has also shown that lasso $L_1$ regularization may effectively avoid over-fitting. In~\citep{Dudik2004}, it is shown that the density estimation in log-linear models using $L_1$-regularized likelihood has sample complexity that grows only logarithmically in the number of features of the log-linear model; \citet{Ng2004} and  \citet{Wainwright2006} show a similar result for $L_1$-regularized logistic regression respectively. 
 
The asymptotic properties of Lasso-type estimates in regression have been studied in detail in~\citep{Knight2000} for a fixed number of variables. Their results say that the regularization parameter $\lambda$ should decay for an increasing number of observations at least as fast as $N^{-1/2}$ to obtain $N^{1/2}$-consistent estimate where $N$ is the number of observations. 
 
\item If $0^0\equiv 0$ is defined, then the $L_0$ regularization contributes a fixed penalty $\alpha_i$ for each weight $w_i\neq 0$. If all $\alpha_i$ are identical, then this is equivalent to setting a limit on the maximum number of non-zero weights. However, the $L_0$ norm is not a convex function, and this tends to make exact optimization of the objective function expensive. 

In general, the $L_q$ norm has parsimonious property (with some components being exactly zero) for $q \leq 1$, while the optimization problem is only convex for $q \geq 1$. So $L_1$ regularizer occupies a unique position, as $q = 1$ is the only value of $q$ such that the optimization problem leads to a sparse solution, while the optimization problem is still convex. 
\end{enumerate}

\section{Structure Learning through Regression} 
Learning a graphical structure by regression is gaining popularity in recent years. The algorithms proposed mainly differ in the choice of the objective loss functions. They are listed in the following according to the different objectives they use. 

\subsection{Likelihood Objective} 
Methods in this category use the negative likelihood or log-likelihood of the data given the parameters of the model as the objective loss function $E_D(\cdot)$.
\begin{itemize} 
\item	In~\citep{Lee2006}, the authors proposed a $L_1$-regularized structure learning algorithm for Markov Random Field, specifically, in the framework of log-linear models. Given a variable set $\mathcal{X} = \{\x_1, \cdots ,\x_n\}$, a log-linear model is defined in terms of a set of feature functions $f_k(\x_k)$, each of which is a function that defines a numerical value for each assignment $\x_k$ to some subset $\x_k \subset \mathcal{X}$ . Given a set of feature functions $F=\{f_k\}$, the parameters of the log-linear model are weights $\theta=\{\theta_k:f_k\in F\}$. The overall distribution is defined as: 

\begin{eqnarray*}
P_{\theta}(\mathbf{x}) = \frac{1}{Z(\theta)}\exp\left(\sum_{f_k\in F}\theta_k f_k(\mathbf{x}) \right)
\end{eqnarray*}

where $Z(\theta)$ is a normalizer called partition function. Given an iid training dataset $\mathcal{D}$, the log-likelihood function is: 

\begin{eqnarray*}
\mathcal{L}(\mathcal{M},\mathcal{D})= \sum_{f_k\in F}\theta_k f_k(\mathcal{D}-M\log Z(\theta)= \theta^{\top}\mathbf{f}(\mathcal{D})-M\log Z(\theta)
\end{eqnarray*}

where $f_k(\mathcal{D})=\sum_{m=1}^M f_k(x_k[m])$ is the sum of the feature values over the entire data set, $\mathbf{f}(\mathcal{D})$ is the vector where all of these aggregate features have been arranged in the same order as the parameter vector, and $\theta^{\top} \mathbf{f}(\mathcal{D})$ is a vector dot-product operation. To get a sparse MAP approximation of the parameters, a Laplacian parameter prior for each feature $f_k$ is introduced such that 
\begin{eqnarray*}
P(\theta_k) = \frac{\beta_k}{2}\exp(-\beta_k |\theta_k|)
\end{eqnarray*}
And finally the objective function is:
\begin{eqnarray*}
\max\limits_{\theta} \quad \theta^{\top}\mathbf{f}(\mathcal{D})-M\log Z(\theta)-\sum_k\beta_k |\theta_k|
\end{eqnarray*}
Before solving this optimization problem to get the parameters, features should be included into the model. Instead of including all features in advance, the authors use \emph{grafting} procedure~\citep{Perkins2003} and \emph{gain-based} method~\citep{Pietra1997} to introduce features into the model incrementally. 

\item In~\citep{Wainwright2006}, the authors restricted to the Ising model, a special family of MRF, defined as 
\begin{eqnarray*}
p(x,\theta) = \exp \left(\sum_{s\in V}\theta_s x_s + \sum_{(s,t)\in E}\theta_{st}x_s x_t -\Psi(\theta) \right)
\end{eqnarray*}
The logistic regression with $L_1$-regularization that minimizing the negative log likelihood is achieved by optimizing: 
\begin{eqnarray*}
\hat{\theta}^{s,\lambda} = \mathop{\arg\min}\limits_{\theta\in \mathbb{R}^p}\left(\frac{1}{n}\sum_{i=1}^n\Big(\log(1+\exp(\theta^{\top}z^{(i,s)}))-x_s^{(i)}\theta^{\top}z^{(i,s)} \Big)+\lambda_n\|\theta_{\setminus s}\|_1 \right)
\end{eqnarray*}

\end{itemize}

\subsection{Dependency Objective} 
Algorithms in this category use linear regression to estimate the Markov blanket of each node in a graph. Each node is considered dependent on nodes with nonzero weights in the regression.
\begin{itemize} 
\item	In~\citep{Meinshausen2006}, the authors used linear regression with $L_1$ regularization to estimate the neighbors of each node in a Gaussian graphical model: 
\begin{eqnarray*}
\hat{\theta}_{i,\lambda}=\mathop{\arg\min}\limits_{\theta:\theta_i=0} \quad \frac{1}{n} \|x_i-\theta^{\top}\mathbf{x}\|_2^2 + \lambda\|\theta\|_1
\end{eqnarray*}
The authors discussed in detail the choice of regularizer weight $\lambda$, for which the cross-validation choice is not the best under certain circumstances. For the solution, the authors proposed an optimal choice of $\lambda$ under certain assumptions with full proof. 

\item	In~\citep{Fan2006}, the authors proposed to learn GGM from directed graphical models using modified Lasso regression, which seems a promising method. The algorithm is listed here in detail. 

Given a GGM with variables $\mathbf{x}=[x_1,\cdots,x_p]^{\top}$ and the multivariate Gaussian distribution with covariance matrix $\Sigma$: 
\begin{eqnarray*}
P(\mathbf{x}) = \frac{1}{(2\pi)^{p/2}|\Sigma|^{1/2}}\exp\left(-\frac{1}{2}\mathbf{x}^{\top}\Sigma^{-1}\mathbf{x} \right)
\end{eqnarray*}

This joint probability can always be decomposed into the product of multiple conditional probabilities: 
\begin{eqnarray*}
P(\mathbf{x})=\prod_{i=1}^p P(x_i|x_{i+1,\cdots,p})
\end{eqnarray*}
Since the joint probability in the GGM is a multivariate Gaussian distribution, each conditional probability also follows Gaussian distribution. This implies that for any GGM there is at least one DAG with the same joint distribution. 

Suppose that for a DAG there is a specific ordering of variables as $1,2,\cdots,p$. Each variable $x_i$ only has parents with indices larger than $i$. Let $\mathbf{\beta}$ denote the regression coefficients and $D$ denote the data. The posterior probability given the DAG parameter $\beta$ is 
\begin{eqnarray*}
P(D|\beta)=\prod_{i=1}^p P(x_i|\mathbf{x}_{(i+1):p},\mathbf{\beta})
\end{eqnarray*}
Suppose linear regression $x_i = \sum_{j=i+1}^p \beta_{ji}x_j+\epsilon_i$ where the error $\epsilon_i$ follows normal distribution $\epsilon_i\sim N(0,\psi_i)$, then 
\begin{eqnarray*}
\mathbf{x} &=& \Gamma\mathbf{x}+\epsilon\\
\epsilon & \sim & N_p(0,\Psi)
\end{eqnarray*}
Where $\Gamma$ is an upper triangular matrix, $\Gamma_{ij}=\beta_{ji},i<j$, $\epsilon=(\epsilon_1,\cdots,\epsilon_p)^{\top}$ and $\Psi = diag(\psi_1,\cdots, \psi_p)$. Thus 
\begin{eqnarray*}
\mathbf{x}=(I-\Gamma)^{-1}\epsilon
\end{eqnarray*}
 
So $\mathbf{x}$ follows a joint multivariate Gaussian distribution with covariance matrix and precision matrix as: 
\begin{eqnarray*}
\Sigma &=& (I-\Gamma)^{-1}\Psi((I-\Gamma)^{-1})^{\top}\\
\Omega &=& (I-\Gamma)^{\top}\Psi^{-1}(I-\Gamma)
\end{eqnarray*}

Wishart prior is assigned to the precision matrix $\Omega$ such that $\Omega\sim W_p(\delta,T)$ with $\delta$ degrees of freedom and diagonal scale matrix $T = diag(\theta_1,\cdots,\theta_p)$. Each $\theta_i$ is a positive hyper prior and satisfies 
\begin{eqnarray*}
P(\theta_i) = \frac{\lambda}{2}\exp(\frac{-\lambda\theta_i}{2})
\end{eqnarray*}

Let $\beta_i = (\beta_{(i+1)i},\cdots,\beta_{pi})^{\top}$, and $T_i$ represents the sub-matrix of $T$ corresponding to variables $\mathbf{x}_{(i+1):p}$. Then the associated prior for $\beta_i$ is $P(\beta_i|\psi_i,\theta_{(i+1):p})=N_{p-1}(0,T_i\psi_i)$~\citep{Geiger2002}, thus: 
\begin{eqnarray*}
P(\beta_{ji}|\psi_i,\theta_j)=N(0,\theta_j \psi_i)
\end{eqnarray*} 
And the associated prior for $\psi_i$ is 
\begin{eqnarray*}
P(\psi_i^{-1}|\theta_i)=\Gamma\left(\frac{\delta+p-1}{2},\frac{\theta_i^{-1}}{2} \right)
\end{eqnarray*}
where $\Gamma(\cdot)$ is the Gamma distribution. Like in~\citep{Figueiredo2001}, the hyper prior $\theta$ can be integrated out from prior distribution of $\beta_{ji}$ and thus 
\begin{eqnarray*}
P(\beta_{ji}|\psi_i) & = & \int_0^{\infty}P(\beta_{ji}|\psi_i,\theta_j)P(\theta_j)\\
&=& \frac{1}{2}\left(\frac{\lambda}{\psi_i} \right)\exp\left(-\Big(\frac{\lambda}{\psi_i} \Big)^{\frac{1}{2}}|\beta_{ji}| \right)
\end{eqnarray*}
Suppose there are $K$ samples in the data $D$ and $x_{ki}$ is the $i$th variable in the $k$th sample, then 
\begin{eqnarray*}
P(\beta_i|\psi_i,D) &\propto& P(x_i x_{(i+1):p},\beta_i,\psi_i)P(\beta_i|\psi_i)\\
&\propto& \exp\left(\frac{\sum_k (x_{ki}-\sum_{j=i+1}^p\beta_{ji}x_{kj})^2+\sqrt{\lambda\psi_i}\sum_{j=i+1}^p|\beta_{ji}|}{-\psi_i} \right) 
\end{eqnarray*}
and
\begin{eqnarray*}
P(\psi_i^{-1}|\theta_i,\beta_i,D) = \Gamma\left(\frac{\delta+p-i+K}{2},\frac{\theta_i^{-1}+\sum_k(x_{ki}-\sum_{j=i+1}^p \beta_{ji}x_{kj})^2}{2} \right)
\end{eqnarray*}

The MAP estimation of $\beta_i$ is: 
\begin{eqnarray*}
\hat{\beta}_i=\mathop{\arg\min}\sum_k\left(x_{ki}-\sum_{j=i+1}^p\beta_{ji}x_{kj} \right)^2 + \sqrt{\lambda\psi_i}\sum_{j=i+1}^p|\beta_{ji}|
\end{eqnarray*}
$\hat{\beta}_i$ is the solution of a Lasso regression.
 
The authors further proposed a \emph{Feature Vector Machine} (FVM) which is an advance to the the generalized Lasso regression (GLR)~\citep{Roth2004} which incorporates kernels, to learn the structure of undirected graphical models. The optimization problem is: 
\begin{eqnarray*}
& \mathop{\arg\min}\limits_{\beta} & \frac{1}{2}\sum_{p,q}\beta_p \beta_q K(f_p, f_q)\\
& \mbox{s.t.} & \left|\sum_p\beta_pK(f_q,f_p)-K(f_q,y) \right|\leq \frac{\lambda}{2}, \forall q
\end{eqnarray*}
where $K(f_i,f_j)=\phi(f_i)^{\top}\phi(f_j)$ is the kernel function, $\phi(\cdot)$ is the mapping, either linear or non-linear, from original space to a higher dimensional space; $f_k$ is the $k$-th feature vector, and $y$ is the response vector from the training dataset. 
\end{itemize}

\subsection{System-identification Objective} 
Algorithms in this category~\citep{Arkin1998,Gardner2003,Glass1973,Gustafsson2003,McAdams1998} get ideas from network identification by multiple regression (NIR), which is derived from a branch of engineering called system identification~\citep{Ljung1999}, in which a model of the connections and functional relations between elements in a network is inferred from measurements of system dynamics. The whole system is modeled using a differential equation, then regression algorithms are used to fit this equation. This approach has been used to identify gene regulatory networks. Here the key idea of this type of approaches is illustrated by using the algorithm in~\citep{Gustafsson2003} as an example.
 
Near a steady-state point (e.g., when gene expression does not change substantially over time), the nonlinear system of the genes may be approximated to the first order by a linear differential equation as: 
\begin{eqnarray*}
\frac{\mathrm{d}x_i^t}{\mathrm{d}t}=\sum_{j=1}^n w_{ij}x_j^t + \epsilon_i
\end{eqnarray*}
where $x_i^t$ is the expression of gene $i$ at time $t$. The network of the interaction can be inferred by minimizing the residual sum of squares with constraints on the coefficients:  
\begin{eqnarray*}
& \mathop{\arg\min}\limits_{w_{ij}} & \sum_t \left(\sum_{j=1}^n w_{ij}x_j^t-\frac{\mathrm{d}x_i^t}{\mathrm{d}t} \right)^2\\
&\mbox{s.t.} & \sum_{j=1}^n |w_{ij}|\leq \mu_i
\end{eqnarray*}

Note that this is essentially a Lasso regression problem since the constraints added to the Lagrangian is equivalent to $L_1$ regularizers. Finally the adjancency matrix $A$ of the network is identified from the coefficients by 
\begin{eqnarray*}
A_{ij}=
	\begin{cases}
		0 & \text{if $w_{ji}=0$}\\
		1 & \text{otherwise}
	\end{cases}
\end{eqnarray*}
One problem with this approach is when the number of samples is less than the number of variables, the linear equation is undetermined. To solve this problem, \citet{D'haeseleer1999} use non-linear interpolation to generate more data points to make the equation determined; \citet{Yeung2002} use singular value decomposition (SVD) to first decompose the training data, and then constrain the interaction matrix by exploring the sparseness of the interactions. 

\subsection{Precision Matrix Objective} 
In~\citep{Banerjee2006}, the authors proposed a convex optimization algorithm for fitting sparse Gaussian Graphical Model from precision matrix. Given a large-scale empirical dense covariance matrix $S$ of multivariate Gaussian data, the objective is to find a sparse approximation of the precision matrix. Assuming $X$ is the estimate of the precision matrix (the inverse of the variance matrix). The optimization of the penalized maximum likelihood (ML) is: 
\begin{eqnarray*}
\max\limits_{X\succ 0} \quad \log\det(X)-\mathrm{Tr}(SX)-\rho\|X\|_1
\end{eqnarray*}
The problem can be efficiently solved by NesterovÕs method~\citep{Nesterov2005}. 

\subsection{MDL Objective} 
\label{sec:4.2.5}
Methods in this category encode the parameters into the Minimum Description Length (MDL) criterion, and tries to minimize the MDL with respect to the regularization or constraints. 
\begin{itemize}
\item	In~\citep{Schmidt2007}, the authors proposed a structure learning approach which uses the $L_1$ penalized regression with MDL as loss function to find the parents/neighbors for each node, and then apply the score-based search. The first step is the $L_1$ regularized variable selection to find the parents/neighbors set of a node by solving the following equation: 
\begin{eqnarray}
\hat{\theta}_j^{L_1}(U)=\mathop{\arg\min}\limits_{\theta}NLL(j,U,\theta)+\lambda\|\theta\|_1
\label{eq:4.4}
\end{eqnarray}

where $\lambda$ is the regularization parameter for the $L_1$ norm of the parameter vector. $NLL(j, U, \theta)$ is the negative log-likelihood of node $j$ with parents $\pi(j)$ and parameters $\theta$: 
\begin{eqnarray}
MDL(G)=\sum_{j=1}^d NLL(j,\pi_j,\hat{\theta}_j^{mle}+\frac{|\hat{\theta}_j^{mle}|}{2}\log n\\
\label{eq:4.5}
NLL(j,\pi(j),\theta)=-\sum_{i=1}^N \log P(X_{ij}|X_{i,\pi(j)},\theta)
\label{eq:4.6}
\end{eqnarray}
where $N$ is the number of samples in the dataset. 

The $L_1$ regularization will generate a sparse solution with many parameters being zero. The set of variables with non-zero values are set as the parents of each node. This hybrid structure learning algorithm is further discussed in Section~\ref{sec:5.1}. 

In general, this regression method is the same as the likelihood objected approaches, since the term of the description length of model in Equation~\ref{eq:4.5} is incorporated into the regularization term in Equation~\ref{eq:4.4}. 

\item	In~\citep{Guo2006}, the authors proposed an interesting structure learning algorithm for Bayesian Networks, which incorporates parameter estimation, feature selection and variable ordering into one single convex optimization problem, which is essentially a constrained regression problem. The parameters of the Bayesian network and the feature selector variables are encoded in the MDL objective function which is to be minimized. The topological properties of the Bayesian network (antisymmetricity, transitivity and reflexivity) are encoded as constraints to the optimization problem. 
\end{itemize}

\chapter{Hybrid Algorithms and Others} 
\section{Hybrid Algorithms} 
\label{sec:5.1}
Some algorithms perform the structure learning in a hybrid manner to utilize the advantages of constraint-based, score-based or regression-based algorithms. Here we list some of them. 
\begin{itemize}	
\item \emph{Max-min Hill-climbing} (MMHC): In~\citep{Tsamardinos2006}, the authors proposed a Max-min Hill-climbing (MMHC) algorithm for structure learning of Bayesian networks. The MMHC algorithm shares the similar idea as the Sparse Candidate Hill Climbing (SCHC) algorithm. The MMHC algorithm works in two steps. In the first step, the skeleton of the network is learned using a local discovery algorithm called Max-Min Parents and Children (MMPC) to identify the parents and children of each node through the conditional independency test, where the conditional sets are grown in a greedy way. In this process, the Max-Min heuristic is used to select the variables that maximize the minimum association with the target variable relative to the candidate parents and children. In the second step, the greedy hill-climbing search is performed within the constraint of the skeleton learned in the first step. Unlike the SCHC algorithm, MMHC does not impose a maximum in-degree for each node. 
	
\item In~\citep{Schmidt2007}, the authors proposed a structure learning approach which uses the L1 penalized regression to find the parents/neighbors for each node, and then apply the score-based search. The first step is the $L_1$ variable selection to find the parents/neighbors set of a node. The regression algorithm is discussed in Section~\ref{sec:4.2.5}.

After the parent sets of all node are identified, a skeleton of the structure is created using the `OR' strategy~\citep{Meinshausen2006}. This procedure is called L1MB ($L_1$-regularized Markov blanket). 
The L1MB is plugged into structure search (MMHC) or ordering search~\citep{Teyssier2005}. In the application to MMHC, L1MB replaces the Sparse Candidate procedure to identify potential parents. In the application to ordering search in~\citep{Teyssier2005}, given the ordering, L1MB replaces the SC and exhaustive search. 

\end{itemize}

\section{Other Algorithms} 
Besides the structure learning algorithms mentioned before, there are some other approaches. They are listed here. 
\subsection{Clustering Approaches} 
The simplest structure learning method is through clustering. First the similarities of any two variables are estimated, then any two variables with similarity higher than a threshold are connected by an edge~\citep{Lukashin2003}. Here the similarity may take different measures, including correlation~\citep{Eisen1998,Spellman1998,Iyer1999,Alizadeh2000}, Euclidean distance~\citep{Wen1998,Tamayo1999,Tavazoie1999} and mutual information~\citep{Butte2000}. Using hierarchical clustering~\citep{Manning2008}, the hierarchy structure may be presented at different scales. 

\subsection{Boolean Models} 
Some algorithms employed the Boolean models for structure learning in gene regulatory network reconstruction~\citep{Thomas1973,Akutsu1999,Liang1998}. These approaches assume the boolean relationship between regulators and regulated genes, and tried to identify appropriate logic relationship among gene based on the observed gene expression profiles. 

\subsection{Information Theoretic Based Approach} 
In~\citep{Basso2005}, the authors employ the information theoretic approaches for reconstructing gene regulatory network. It first identify pairs of correlated genes based on the measurement of mutual information. It then eliminates indirect interaction among genes by applying the well-known staple of data transimission theory, the ``data processing inequality'' (DPI). There are two things unclear about this approaches: 1) Since the results could be sensitive to the order of elimination, it is important to provide a justification about the order of edges to be eliminated. 2) Most of the discussion within the paper is limited to loops with three edges. It is important to know how to address the cycles with more than three genes. 

\subsection{Matrix Factorization Based Approach} 
Methods in this category use matrix factorization techniques to identify the interactions between variables. The matrix factorization algorithms used includes singular value decomposition~\citep{Alter2000,D'haeseleer1999,Raychaudhuri2000}, max-margin matrix factorization~\citep{DeCoste2006,Rennie2005,Srebro2005} and non-negative matrix factorization~\citep{Badea2005,Paatero1994,Hoyer2004,Lee2001,Shahnaz2006,Weinberger2005}, network component analysis (NCA)~\citep{Liao2003}. Readers are referred to read a technical report~\citep{Jin2006} for more details of this method. 

\bibliographystyle{plainnat}
\bibliography{references}

\end{document}